\documentclass{article}
\pdfoutput=1

\usepackage{arxiv}

\usepackage[utf8]{inputenc} 
\usepackage[T1]{fontenc}    
\usepackage{hyperref}       
\usepackage{url}            
\usepackage{booktabs}       
\usepackage{amsfonts}       
\usepackage{nicefrac}       
\usepackage{microtype}      
\usepackage{amsmath}
\usepackage{cleveref}       
\usepackage{lipsum}         
\usepackage{graphicx}
\usepackage{natbib}
\usepackage{doi}
\usepackage{multirow}

\title{One model Packs Thousands of Items with Recurrent Conditional Query Learning}

\date{}

\author{Dongda Li$^{1}$, Zhaoquan Gu$^{1}$, Yuexuan Wang$^2$$^3$, Changwei Ren$^2$, Francis C.M. Lau$^3$ \\
	$^1$ Guangzhou University, $^2$ Zhejiang University, $^3$The University of Hong Kong \\
	\texttt{lidongda@gzhu.edu.cn, zqgu@gzhu.edu.cn, amywang@zju.edu.cn} \\
	\texttt{rcw@zju.edu.cn, fcmlau@cs.hku.hk}
}

\renewcommand{\headeright}{}
\renewcommand{\undertitle}{}
\renewcommand{\shorttitle}{One model packs thousands of items with Recurrent Conditional Query Learning}

\begin{document}

\maketitle

\begin{abstract}
Recent studies have revealed that
neural combinatorial
optimization (NCO) has advantages over conventional algorithms in many
combinatorial optimization problems such as routing, but it is less
efficient for more complicated optimization tasks such as packing which
involves mutually conditioned action spaces. In this paper, we propose a
Recurrent Conditional Query Learning (RCQL) method to solve both 2D and
3D packing problems.
We first embed states by a recurrent encoder,
and then adopt attention with conditional queries from previous actions.
The conditional query mechanism fills the information gap between learning steps, which
shapes the problem as a Markov decision process. Benefiting from the
recurrence, a single RCQL model is capable of handling different sizes of packing
problems.
Experiment results show that RCQL can effectively
learn strong heuristics
for offline and online strip packing problems (SPPs), outperforming a wide
range of baselines in space utilization ratio. RCQL reduces the average bin gap ratio by 1.83\% in offline
2D 40-box cases and 7.84\% in 3D cases compared with state-of-the-art methods.
Meanwhile, our method also achieves 5.64\% higher space utilization ratio
for SPPs with 1000 items than the state of the art.

\end{abstract}


\section{Introduction}

How to pack boxes into the smallest bin? The answer to this deceivingly simple
question is actually one of the most crucial ones that power today's
e-commerce. Zillions of boxes
are being dispatched to customers each day from suppliers far and near.
Being able to efficiently pack boxes compactly into bins and containers
translates directly into reduced shipping costs and energy.

Packing problems, which are a type of classical combinatorial
optimization problem (COP), have been extensively studied for many
decades~\citep{christensen2017approximation} by researchers in
operational research as well as those in computer science.
Such a problem is strongly NP-hard~\citep{martello2000three} even when
only one bin is considered,
which is called the strip packing problem (SPP).
Each packing step involves three
interrelated actions, namely, item selection, rotation and positioning;
what makes the problem acutely difficult is that
later actions could be strongly affected by the previous actions.
Therefore, exact algorithms~\citep{silva2019exact} to achieve the optimal
solution expectedly require a huge amount of computation time, and so
they are only suitable for small-scale problems.

Other than exact algorithms, there are approximation and heuristic
methods which tend to run faster.
Some approximation algorithms~\citep{christensen2017approximation}
provide solutions with a lower bound in polynomial time.
Unfortunately, the results of these algorithms turned out
to be not even as good as some simple
packing algorithms~\citep{crainic2008extreme}.
Heuristic approaches~\citep{baltacioglu2006distributor,gonccalves2011parallel,egeblad2009heuristic,duong2015heuristics}
define some rules based on human experience and knowledge about the
problem. Some recent works~\citep{wei2012reference,wei2017improved} tried to
improve on previous proposed heuristics by adding more rules;
these methods however require substantial prior knowledge of
the problem and thus lack flexibility in terms of the problem setting.
Meta-heuristics~\citep{rakotonirainy2020improved,zeng2016iterated,mostaghimi2017three,wu2010three} apply search methods to try to find
better solutions in an iterative manner, but such a process takes a very
long time at arrive at a good solution.

In fact, the explicit or hand-crafted rules of any special setting of a
problem can be interpreted as policies in making decisions.
Moreover, policies can be modeled by neural networks (NNs) in reinforcement learning.
Hence, many recent studies~\citep{bello2016neural,nazari2018reinforcement,lu2019learning}
have adopted NNs and reinforcement learning to solve
classic COPs including the Traveling Salesman
Problem (TSP), the Vehicle Routing Problem (VRP), etc.
There were also some learning-based
attempts~\citep{duan2019multi,cai2019reinforcement}
which utilize reinforcement learning with neural network models to
solve packing problems.

However, these existing approaches either pay no attention to the
connection between interrelated actions or they rely on hand-crafted rules
in the learning algorithm.
Without exploiting the connection between interrelated actions, the
reinforcement
learning environment becomes a partially observable Markov Decision Process
(MDP)~\citep{sutton2018reinforcement} which is hard to generalize as a
good policy due to lack of information. Those methods that use
hand-crafted rules are not only unlikely to achieve an optimal
solution, but are also overly sensitive to problem settings. Besides their
relatively poor performance, previous learning based
approaches also suffer from upsurge of memory costs as the number of boxes
to be packed increases.
These methods simply take all previously-packed boxes and the
current candidate box states to infer the next packing
actions, so the memory overhead increases rapidly as the problem size grows.
Even worse, these models have to be re-trained for problems with
different sizes because the model structure is problem size-related,
thus posing a great challenge for practical applications.

In order to overcome the problems with previous learning methods,
we introduce an end-to-end learning method called Recurrent Conditional Query Learning (RCQL) that
directly addresses the information gap between interrelated actions.
With the inner conditional query mechanism, the learning process is similar
to a fully observable MDP, which makes training by reinforcement learning much easier.
Compared with models that output several interrelated actions
simultaneously where the action set per step is formed as a Cartesian
product of the sub-action spaces, the action space of the
RCQL model in each forward step equals the corresponding sub-action
space. Such smaller action space allows us to apply a simpler model to
fit the learning policy, which makes the model easier to train and more
memory-efficient. Unlike previous learning-based approaches~\citep{duan2019multi},
RCQL does not involve any hand-craft rules or heuristics,
and the whole model are trained by gradient-based learning method,
making it an end-to-end learning method.

More specifically, the packing problem requires three mutually conditioned
sub-actions: box selection, rotation and positioning.
To fill the gap between sub-actions, we adopt the RCQL model as follows.
First of all, the packing problem is formulated as an MDP for applying
reinforcement learning. Then the previous sub-actions are embedded as a
attention in the next sub-action decoder. After all three sub-actions
have been generated, a packing step is performed and the observations are updated.
Finally, we adopt the
entropy regularized actor-critic algorithm~\citep{haarnoja2018soft}
to update the model parameters. In addition, because unpacked boxes and
packed boxes have different properties, the input state of RCQL
model is divided into two substates, which ensures that the model
has a targeted representation of each substate. Furthermore, in order to deal with the
sparse reward problem~\citep{andrychowicz2017hindsight} in the packing
problem, we design a novel gap-reward function, which entices the
learning to converge faster.

Apart from the conditional query mechanism, the RCQL model also
incorporates some recurrent attention features. That is, the model
not only acquires information from the current environment state,
but also gets information via an attention mechanism from previous
hidden states to infer the subsequent step actions.
In order to make recurrent attention work, the dynamic state updating is
devised. Combining dynamic state updating and recurrence, the
recurrence attention model enables learning dependency beyond current
inputs without disrupting temporal coherence. This ensures that a
fixed-size model can solve problems of different sizes without
undermining the performance. At the same time, the model is suitable for
on-line packing problems that are generally too hard for conventional
algorithms.

We conduct extensive experiments to evaluate different models and the
results show that the RCQL model
achieves a lower gap ratio in both 2D and 3D packing than   heuristic methods and existing learning approaches.
Specifically, our model improves the space utilization ratio in 3D
packing (40 boxes) by 29.32\% compared with genetic algorithms, and
reduces the bin gap ratio in almost every case by more than 7\% compared with
the state-of-the-art learning approaches. Furthermore, RCQL can scale to
handle even thousands of boxes with only 2.16M model parameters, thanks to the
recurrence feature. In addition, our method also achieves superior
results for on-line packing problems.

The contributions of this paper are as follows:
\begin{enumerate}
\item We propose the
first end-to-end learning method that solves the SPP
in both offline and online mode, and under both 2D and 3D settings.
\item We formulate the
packing problem as an Markov decision process (MDP), and design a general packing environment
tailored to reinforcement learning algorithms.
\item We propose a recurrent
conditional query learning model to address large-scale strip packing
problems.
\item We conduct extensive experiments on large-scale packing
problems and the results show that our model outperforms
state-of-the-art methods.
\end{enumerate}

The rest of the paper is organized as follows. We introduce packing
problems and conventional solution approaches in the next section. Deep
reinforcement learning and its application in COPs are discussed in
Section~\ref{prel}. The MDP formulation of the packing problem and the
design of the RCQL
model are presented in Section~\ref{MDP} and Section~\ref{model},
respectively. We test the RCQL model and present the comparison
results in Section~\ref{experiments}. Section~\ref{conclusion}
concludes the paper.

\section{Background}
\label{back}

In this section, we introduce packing problems and conventional
approaches for solving them.

\subsection{Packing Problems}

Packing problems first appeared as
a class of optimization problem in mathematics, that
involve packing objects into containers. The goal is to either
pack objects into a single bin as compactly as possible or pack all
objects into as few bins as possible.

\paragraph{Strip packing problem}
For simplicity, and as our first
goal, we have a number of boxes and we want to pack them into minimal space in the bin. Specifically, we have a fixed bottom
size rectangle (2D) or a cubic (3D) bin, and the object is to minimize
the final height of the bin and thus achieve a higher space utilization
ratio. The problem is an SPP with
rotations~\citep{christensen2017approximation}, which is a subclass of
geometric packing problems, and we follow the definition of this problem
as in \citep{wu2010three}.

\paragraph{Online and offline packing} Reflecting many practical
situations, packing problems are divided into two categories, namely,
offline packing and online packing. In offline packing, all candidate
boxes are given in advance, so the packing algorithm can choose the most
suitable one to pack in each step. In online packing, the
candidate boxes are given one by one, which means the algorithm can only
pack the given box to the bin at every step.

\paragraph{Packing procedure} With respect to offline packing, each step comprises three sub-actions: 1)Selecting a target box from all unpacked boxes.
2) Choosing the rotation of the selected box.
3) Outputting the position of the rotated box relative to the bin.

These three sub-actions are ordered and mutually conditioned. For
the online variant, the box selection step is skipped.
In this paper, both offline and online packing problems are
addressed in the 2D and 3D cases. We elaborate our formulation in the 3D
offline case unless specifically stated otherwise in following sections.

\subsection{Conventional approaches for packing problems}
Packing has been intensively studied in the last few decades. Three
types of methods are employed in previous works, namely, exact,
approximation and heuristic algorithms.
The current best exact algorithm\citep{silva2019exact} takes hours to
solve the packing problem of only 12 items, which is obviously not
efficient enough for practical use.
Some approximation
algorithms~\citep{christensen2017approximation} provide a guarantee
for the quality of the solution via worst-case bound. However,
state-of-the-art approximation algorithms only offer solutions
that are very close in terms of solution quality to simple
or sometimes even inefficient heuristic
algorithms~\citep{duong2015heuristics}. Heuristic approaches define some rules to
pack items. Although heuristics cannot guarantee the quality of solution
achieved, in practice, they are better than approximation
algorithms. Meta-heuristics further improve on the heuristic solutions by
applying search-based techniques such as hill climbing,
tabu search~\citep{zeng2016iterated}, simulated
annealing~\citep{rakotonirainy2020improved}, genetic
algorithms~\citep{wu2010three} etc., but they tend to require more computing
time.

We can see that, in recent years, research on general packing problems has
made little progress. Many recent studies have turned to packing problems with
specific constraints~\citep{baldi2019generalized, ding2019rectangle,
grange2018algorithms,martinez2017matheuristics}. It is hard to improve
general packing results in traditional ways.

\section{Deep Reinforcement Learning and Its Applications in COPs}
\label{prel}


Neural Networks (NNs) and deep reinforcement learning have enjoyed rapid
development in recent years. Since heuristic policies can be
parameterized using NNs, much
recent studies have adopt this promising technique to solve COPs,
especially routing problems~\citep{bello2016neural,
nazari2018reinforcement, kool2018attention}. At the same time, COPs such
as routing and packing are strongly NP-hard, so it is unrealistic to get
optimal solutions to use
as labels within an acceptable time range. With reinforcement
learning, the agent improves the policy based on its own experience,
which is suitable for the unlabeled problem.

\subsection{Reinforcement Learning}
Reinforcement learning problems can be formulated as Markov decision processes (MDPs),
in which precise theoretical statements can be made. COPs can be defined as policy search in an MDP which
comprises a state space $\mathcal{S}$, an action space $\mathcal{A}$,
and a reward function $r: \mathcal{S} \times \mathcal{A} \to
\mathbb{R}$. For infinite horizon problems, a discount factor $\gamma
\in (0,1]$ is also included.
A policy $\pi(a_t|s_t)$ is used to select actions in the MDP. The goal
of reinforcement learning is to find a $\pi^*(a_t|s_t)$ that maximizes
the cumulative discounted reward from the start state $s_1$, denoted by the
objective function $ J(\pi)=\mathbb{E}[\sum_{k=1}^{\infty}\gamma^{k-1}
r(s_k,a_k)|\pi]$, where $\gamma$ is a discount factor determining the
priority of short-term rewards.

\paragraph{Soft Actor-Critic}

Soft Actor-Critic~\citep{haarnoja2018soft} (SAC) is a state-of-art
reinforcement learning algorithm, which learns a policy
$\pi_{\theta}(a|s)$ and critic $Q_{\phi}(s,a)$ by maximizing a weighted
objective of reward and policy entropy,
$\mathbb{E}_{s_t,a_t\sim\pi}[\sum_t r_t + \alpha
\mathcal{H}(\pi(\cdot|s_t))]$. Here, $\alpha$ is the temperature
parameter that determines the relative importance of the entropy term
versus the reward, and $\mathcal{H}$ is
the entropy of actions produced by policy $\pi$. In the actor-critic
algorithm, the policy objective function is
$\mathbb{E}_{\pi_\theta}[Q_\phi(s,a)\log \pi_\theta(s,a)]$. In more
advanced settings, the Generalized Advantage Estimation
(GAE)~\citep{schulman2015high} is adopted to reduce the Q value estimation
error. As a result, the policy objective function becomes
$\mathbb{E}_{\pi_\theta}[A(s,a)\log \pi_\theta(s,a)]$, where $A(s,a)$ is
the action advantage.

\subsection{Deep learning models for COPs}

Since both the input and output of most COPs are sequences, people borrow
the sequence to sequence (seq2seq) models to solve COPs. Seq2seq models
are originally used in neural machine translation
(NMT)~\citep{bahdanau2014neural, vinyals2015order,luong2015effective,
vaswani2017attention, shen2018bidirectional}. The most commonly used
seq2seq models are RNN~\citep{wang2016text} and the attention
model~\citep{vaswani2017attention}. RNN is well known, so here we briefly
introduce the attention model.

\paragraph{Attention} Attention was first proposed in the NMT
work~\citep{bahdanau2014neural}, which weights the RNN hidden vector of
each input and combines it with previous outputs to infer the current
output. The transformer~\citep{vaswani2017attention} further improves the
attention mechanism by using the self-attention module. A self-attention
module computes the representation at a position in a sequence by
attending to all positions and taking their weighted average in an
embedding space, which allows NNs to focus on different parts of their
input.

\subsection{Previous work on NCO and its limitations}

By combining reinforcement learning and seq2seq models, many neural combinatorial
optimization (NCO)
studies~\citep{mazyavkina2021reinforcement} have improved the results for
COPs, especially for
routing problems~\citep{bello2016neural, nazari2018reinforcement,
kool2018attention}. Pointer Networks~\citep{vinyals2015pointer} adopt
attention as a pointer to select a member of the input sequence as the
output. \citep{bello2016neural} and ~\citep{nazari2018reinforcement} view
TSP and VRP as MDP, and they both apply a policy gradient
algorithm to train their models. In \citep{kool2018attention}, the result
of the routing problem is further improved by using the attention
model~\citep{vaswani2017attention}.

There are a few attempts using NCO
to solve packing problems. In~\citep{cai2019reinforcement}, Reinforcement
learning is adopted to get some packing results as initialization to
accelerate the original heuristic algorithms. The authors
of~\citep{duan2019multi} propose a selected learning approach to solve 3D
flexible bin packing problems by balancing the sequence and orientation
complexity. They adopt pointer networks~\citep{vinyals2015pointer} with
reinforcement learning to obtain the box selection and rotation decisions,
and apply a greedy algorithm to select the box location, which is not an
end-to-end learning method and does not perform better than greedy
algorithms. More importantly, this hybrid method does not view the
entire packing problem as one complete optimization process, and therefore
different algorithms of sub-actions may have conflicting goals in the
optimization process, which leads to inferior results.



The key difference between the routing problem and the packing problem
is that each step of the routing problem only needs to select one item
from the input sequence, whereas the packing problem requires three
sub-actions to pack a box into the bin. This latter kind of action space is
called parameterized action space~\citep{masson2016reinforcement,
fan2019hybrid, neunert2020continuous} in reinforcement learning, which
requires the agent to select multiple types of actions in each action
step. When the action space contains finite actions each parametrized by
a continuous parameter, it is called a hybrid action space.

To handle reinforcement learning problems with parameterized action space,
Hausknecht et al.~\citep{hausknecht2015deep} extend deep deterministic
policy gradients (DDPG)~\citep{silver2014deterministic} to parameterized
action space. But their approach suffers from the $tanh$ saturation
problem~\citep{xu2016revise} in continuous action space.
Q-PAMDP~\citep{masson2016reinforcement} alternates between learning action
selection and parameter selection polices to make the training process
more stable. But the model still outputs all the parameters in one
forward pass. As a simple example, if there are three parameters, and
each parameter has a discrete action space of size 10, then the model
should output $10^3$ actions in each forward step.
So
the performance of Q-PAMDP is limited by the action space size of the parameter policy.
The authors of~\citep{wei2018hierarchical} propose a hierarchical
approach, by which they condition the parameter policy on the output of
the discrete action policy, and apply a reparameterization
trick~\citep{kingma2015variational} to make the model differentiable.

Nevertheless, all these studies treat each sub-action as part of one
forward-pass result of the policy model. Thus, the number of outputs of
the model will be the Cartesian product of the candidate output of each
sub-action, which significantly increases the output options of the
model, making the model hard to generalize and to learn to produce a
good solution. A more reasonable approach is to update the sub-actions
separately with different models.

Different from existing approaches, our method simply embeds the
previous actions as an attention query for the model to reason for the
subsequent actions, and each sub-action has its own model branch. After
a full packing step is finished, we perform one-step optimization along
every sub-action model branch. In this way, we reduce the output size of
the learning model in each step, which reduces the size of the model and
improves its learning performance.

\section{Formulating Packing Problem as an MDP}
\label{MDP}

In this section, we formulate the packing problem as an MDP and prepare
it for reinforcement learning. In particular, we introduce the state,
action and reward construction of our reinforcement learning-based
method.

\subsection{Dynamic State Updating}

For a large scale packing problem, it is impractical to
view all environmental information as state input due to the large memory
cost. Thus, we treat the problem as an infinite horizon problem. The reason why
this is still rational is that the learning algorithm is only required to
learn a packing pattern based on the current upper surface of the bin
and the candidate boxes.

In the offline packing problem, the unpacked boxes are used as
candidates to select the next box, while the arrangement of
packed boxes are used to infer the location of the next box. As a result, the state of
the packing problem consists of two substates: the packed box substate
$\pmb{s}_p$ and the unpacked box substate $\pmb{s}_u$, where
$\pmb{s}_p=\{s_{p,1},s_{p,2},\ldots,s_{p,n_p}\}$, and
$\pmb{s}_u=\{s_{u,1},s_{u,2},\ldots,s_{u,n_u}\}$. Here, $n_p$ and $n_u$ are
the sizes of packed box substate and unpacked box substate respectively,
which is defined as the context size.
$s_{u,i}=\{l_i,w_i,h_i\}$, where $l_i$, $w_i$ and $h_i$ denote the
length, width and height of box $i$, respectively.
$(x_i,y_i,z_i)$ denotes the
left-front-bottom corner of the $i$-th box placed inside the bin. Thus,
the $i$-th packed state is represented as $s_{p,i} = \{l_i', w_i', h_i',
x_i, y_i, z_i\}$, where $(l_i',w_i',h_i')$ are the rotated box
length, width and height respectively of the box $i$.
By dividing the input state into two substates with different
attributes, the model can provide specialized representations of
these substates, thereby improving packing results.

\begin{figure}[tbhp]
\centering
\includegraphics[width=.8\columnwidth]{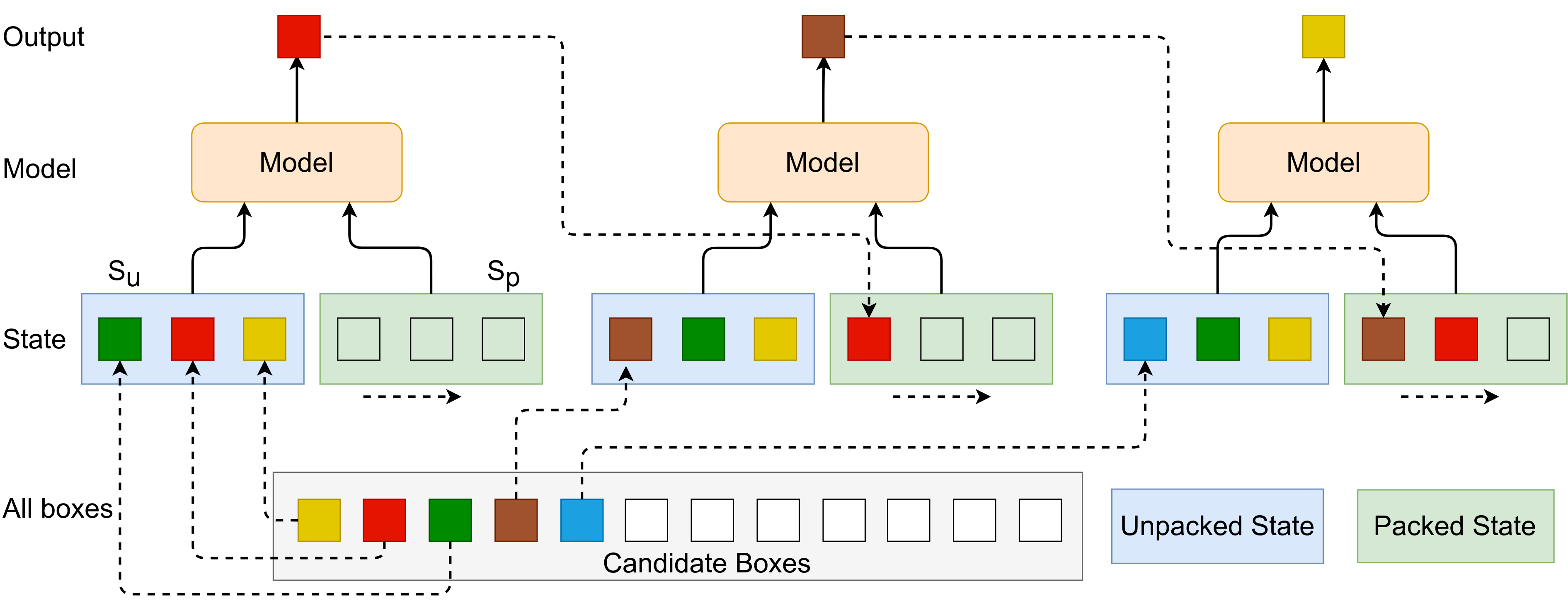}
\caption{Illustration of dynamic state updating with context size 3. The
dashed line represents the data copy flow, and the solid line represents
the model forward-pass process. The model here serves as a reinforcement
learning policy, which takes input states and returns actions. Best viewed in color.}
\label{fig:state_update}

\end{figure}

As shown in Fig.~\ref{fig:state_update}, after every packing step, the
packed box $i$ in $\pmb{s}_u$ is discarded and replaced with a new box. In
the mean time, the packed box state $s_{p,i}$ is pushed to the First In
First Out (FIFO) queue $\pmb{s}_p$ which keeps the state context fresh.
Although state $\{\pmb{s}_p, \pmb{s}_u\}$ seems enough for viewing a whole
packing step as an MDP, it is not sufficient for the last two
sub-actions of a packing step due to the mutually conditioned feature of
sub-actions. In the rotation step, the state should be $\{\pmb{s}_p,
s_c\}$, where $s_c =\{l_c,w_c,h_c\}$ is the current packing box size,
and for the positioning step, $s=\{\pmb{s}_p, s_c'\}$, where
$s_c'=\{l_c',w_c',h_c'\}$.

\subsection{Action Space} In a packing problem, each sub-action has its
own action space. In the box selection step, the action chooses a
box from the current unpacked context, so the action is selected from 1
to $n_u$. In the box rotating step, there are 6 options for the current
selected box, and in the positioning step, each side of the bottom bin
is discretized into $n_s$ slots, so the action space is $n_s\times
n_s$.

\subsection{Reward Function}

Unlike supervised learning, the agent in reinforcement learning learns
from experience and tries to find the policy that can get more accumulative
rewards. Therefore, to design a proper reward signal is crucial for
producing a good solution. In a packing problem, the goal is
to minimize the height of the bin with a given width and length. The
most straightforward way is to adopt the negative change of bin height
$-\Delta h=h_{t}-h_{t+1}$ as the reward signal in every packing step.
However, because not every packaging step will cause the height of the bin to change,
this leads to sparse rewards, which is one of the biggest
challenges~\citep{andrychowicz2017hindsight} in reinforcement learning.

\begin{align}
\label{reward_orignal}
\begin{split}
g_t &= W L H_t-\sum_{i=1}^t(w_i l_i h_i) \\
\tilde{r}(s_t,a_t) &= \Delta g_t = g_{t-1}-g_t
\end{split}
\end{align}

To address the sparse reward problem, we design the reward signal based
on the change of the current volume gap of the bin.
As shown in (\ref{reward_orignal}), the volume gap $g_t$ of the
packing step $t$ is defined as the current bin volume minus the total
volume of packed boxes, where $W,L,H$ are the width, length and height of
the bin, respectively. The reward $\tilde{r}(s_t,a_t)$ of the packing
step is defined as $\Delta g_t$, so the accumulated reward becomes the
final gap of the bin, which is linearly proportional
to the negative final bin height
$-H_n$ as formulated in (\ref{accum_reward}). By doing so, the agent
always gets a meaningful reward signal even when there is no increase in total bin height.

\begin{align}
\label{accum_reward}
\begin{split}
R &= \sum_{t=1}^n \tilde{r}(s_t,a_t) =-W L H_n+\sum_{i=1}^n(w_i l_i h_i)
\end{split}\\
\label{reward}
r(s_t, a_t) &= \tilde{r}(s_t,a_t) +\alpha \mathcal{H}(\pi(\cdot|s_t))
\end{align}

As described before, the problem is considered an infinite horizon
learning problem, and a discount factor $\gamma$ is adopted to avoid $R$
becoming infinite. In addition, in order to encourage exploration, an
entropy maximization strategy is also incorporated. Eventually, the
reward function of each packing step is formulated as (\ref{reward}), similar to SAC~\citep{haarnoja2018soft}.

\section{Recurrent Conditional Query Learning}
\label{model}

In this section, we introduce the RCQL model. In order to enable the
model to solve large-scale optimization problems with mutually conditioned
actions, two modules, namely, the recurrent attention encoder and the
conditional query decoder, are incorporated into the model.

\subsection{Recurrent Attention Encoder}
\label{recurrent}

The Transformer architecture~\citep{vaswani2017attention} has been shown to excel in many
seq2seq tasks, because the self-attention layer of Transformer enables the network
to capture contextual information from the entire input sequence, thereby providing
the relationship between the features of different input data.
However, the computational and memory costs of such a network grow
quadratically with the sequence length~\citep{child2019generating} and
thus it is hard to apply this method to long sequences.

As mentioned in Section~\ref{MDP}, the packed state $\pmb{s}_p$ is
saved in a FIFO to keep the context fresh. Although this dynamic state
updating mechanism leads to smaller memory costs, for large scale
packing problems, there is still a trade-off between memory cost and
context length. That is, the model will overlook long-term dependencies
when the context size is too small. The model simply cannot obtain the
information of the state of packed boxes that are out of context.
On the other hand, with a large context size the memory cost could
greatly increase.

\begin{align}
\begin{split}
\label{eq:recurrence}
\tilde{\pmb{h}}_{\omega+1}^{n-1} &=\textit{Concat}[\pmb{h}_{\omega}^{n-1}, \pmb{h}_{\omega+1}^{n-1}] \\
\pmb{q}_{\omega+1}^n &= \pmb{W}_q^T \pmb{h}_{\omega+1}^{n-1}, \\
\pmb{k}_{\omega+1}^n &= \pmb{W}_k^T \tilde{\pmb{h}}_{\omega+1}^{n-1}, \\
\pmb{v}_{\omega+1}^n &= \pmb{W}_v^T \tilde{\pmb{h}}_{\omega+1}^{n-1} \\
\end{split}
\end{align}

\begin{equation}
\label{eq:att}
\pmb{h}_{\omega+1}^{n} = soft\max(\pmb{q}_{\omega+1}^n {\pmb{k}_{\omega+1}^n}^T)\pmb{v}_{\omega+1}^n
\end{equation}

To address the conflict between long-term dependence and memory cost,
a recurrence feature is added to the
Transformer layer to encode the packed
state. In every transformer self-attention layer, a context cache of the
previous hidden state $\pmb{h}_{\omega}^{n-1}$ is concatenated with the
current hidden state $\pmb{h}_{\omega+1}^{n-1}$, where $\omega$ denotes
the context number, and $n$ denotes the layer number. Note that this
recurrence is different from that of Transformer-XL~\citep{dai2019transformer}.
Here the context only moves one item forward per step since only one new
packed box is available for the packed state in a packing step, so the
previous hidden state $\pmb{h}_{\omega}^{n-1}$ is also treated as FIFO.
The recurrent attention encoder layer is produced by
(\ref{eq:recurrence}), where $\pmb{q}, \pmb{k}, \pmb{v}$ denote query,
key and value respectively in Transformer, and
$\pmb{W}_q, \pmb{W}_k, \pmb{W}_v$ are the trainable parameters for
$\pmb{q}, \pmb{k}, \pmb{v}$, respectively. In practice, one could extend
single-head attention in~(\ref{eq:att}) to Multi-Head
Attention~\citep{vaswani2017attention} (MHA) to capture a mixture of
affinities. MHA computes $n_h$ single-head attentions in parallel, and
then concatenates them along the feature dimension.

\begin{figure}[tbhp]

    \centering
    \includegraphics[width=0.8\columnwidth]{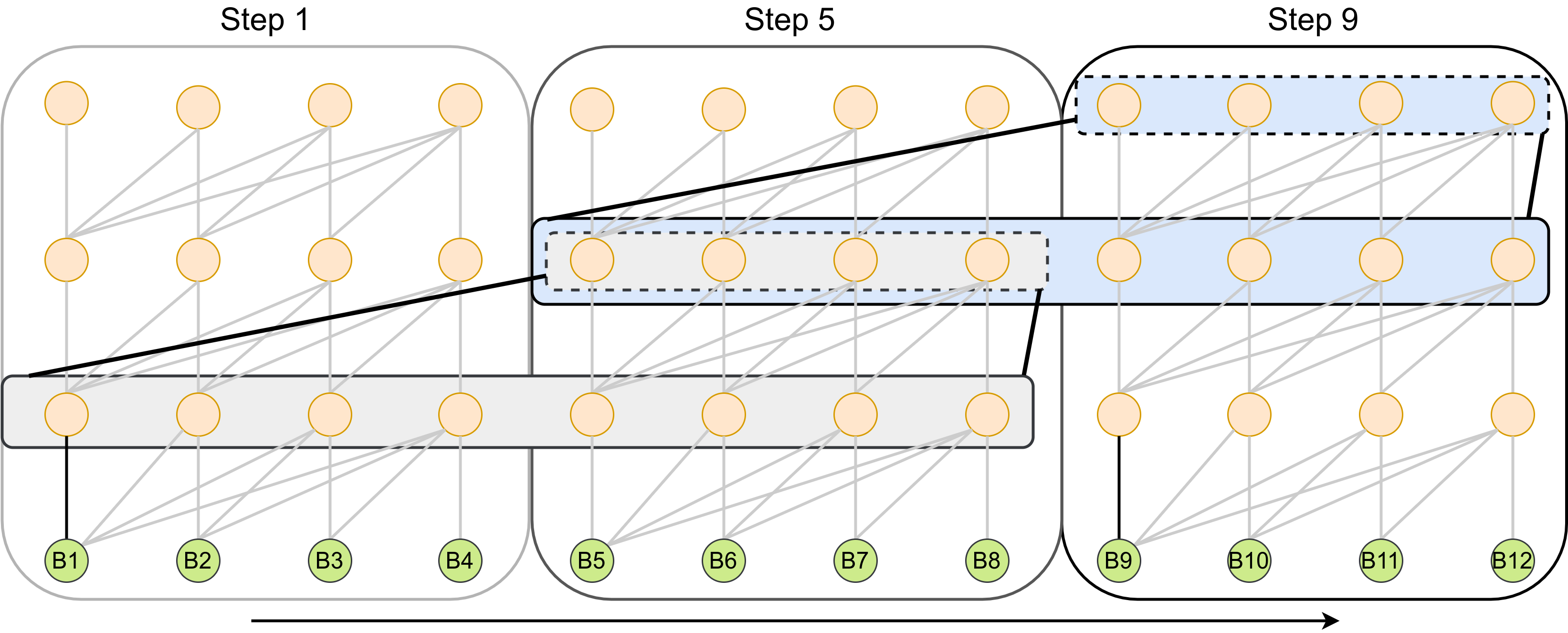}

    \caption{Demonstrating 3 layers of recurrent attention with a context
    size of 4. The solid lines joining layers show the information
    dependence flows in step 9, and the dotted line block is the attention
     result of the previous layer.}
    \label{fig:recurrence}

\end{figure}

As shown in Fig.~\ref{fig:recurrence}, by using the recurrence feature,
the model can get the information of the previous $L-1$ context blocks,
where $L$ denotes the layer number of the encoder. As a result, the
recurrent attention encoder is able to achieve longer dependence with a
smaller model size. Consequently, it is easy to design a suitable model
that can receive useful information of all packed boxes visible on the upper surface of
the bin when packing a new one. Besides, the model does not have to
apply positional encoding as the packed state already contains position
information that includes the sizes and coordinates of the packed boxes.

\subsection{Conditional Query Decoder}

\begin{figure}[tbhp]

\centering
\includegraphics[width=0.6\columnwidth]{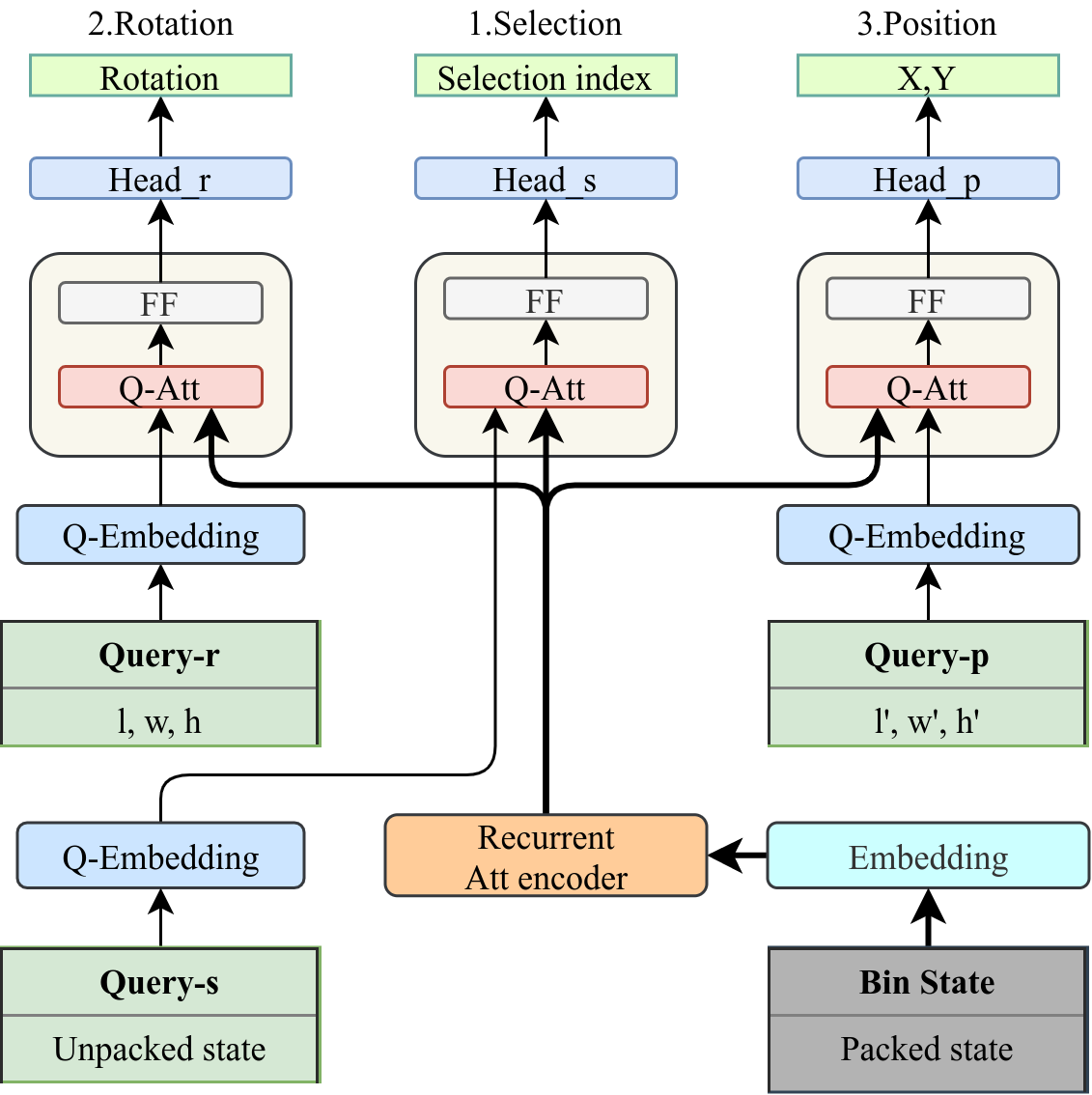}
\caption{Recurrent Conditional Query Model.}
\label{fig:decoder}

\end{figure}

After encoding the packed state into an embedded vector $\pmb{h}_e^n$ by
the recurrent attention encoder, we design the conditional query
mechanism to handle the connection between sub-actions. As shown in
Fig.~\ref{fig:decoder}, for packing problems, the model performs three
sub-actions sequentially, namely, box selection, rotation and
positioning for the selected box, and each sub-action has one
independent decoder.

These three decoders have the same basic MHA structure, but their
embedded conditional query signals that are from previous steps are
different. They also share the same attention key and value as calculated
by the recurrent attention encoder, which embed the information from
the packed state. A conditional query layer contains a Queried MHA
(Q-Att) and a
Feed-Forward (FF) layer. After the conditional query layer, each decoder
has several linear head layers to project the conditional query outputs
$\pmb{q}_{out}$ to each sub-action probability.

\begin{align}
\begin{split}
\label{eq:query}
\pmb{q}_s &= \pmb{W}_{qs}^T (\pmb{W}_{s} \pmb{s}_u), \;
\pmb{q}_r = \pmb{W}_{qr}^T (\pmb{W}_{r} s_c), \;
\pmb{q}_p = \pmb{W}_{qp}^T (\pmb{W}_{p} s_c') \\
\pmb{q}_{o} &= \text{MHA}(\pmb{q}, \pmb{W}_k^T \pmb{h}_{e}^n, \pmb{W}_v^T \pmb{h}_{e}^n)
\end{split}
\raisetag{12pt}
\end{align}

To be more specific, as formulated in (\ref{eq:query}), the unpacked
state $\pmb{s}_u$ is embedded as a query vector $\pmb{q}_s$ in the box
selection step. After decoding, the candidate box is selected from boxes
in the unpacked state, in which
$\pmb{W}_{s}, \pmb{W}_{r}, \pmb{W}_{p}$ are weight matrices of three
sub-actions for query embedding. Then in the rotation step, the
conditional query $\pmb{q}_r$ is constructed by picking up the selected
box shape information $s_c$ from the unpacked state and embedding it in
a linear layer. The model calculates the position of the selected box
relative to the bin in the final step. To incorporate all previous results,
the positioning step query $s_c'$ combines the rotation of the selected
box.

Throughout the entire forward pass of packing one box, the data stream
passes through the encoder and each decoder once. The target query is
extracted from the candidate box state based on the previous outputs.
Via the conditional query, the model receives information from the hidden
vectors of the encoder as well as previous sub-action outputs, which
ensures that every sub-action decoding step is an MDP.

\subsection{Training}

As mentioned earlier, we view the packing process as an MDP and apply
reinforcement learning, or specifically, the actor-critic
algorithm~\citep{konda2000actor}. The model presented earlier is an actor
model. Here we briefly describe the critic model, the actor policy and
the training process.

\subsubsection{The Critic Model}
The critic structure is similar to the box selection model of the actor model,
which consists of a recurrent attention encoder and a conditional
decoder with the unpacked state as input followed by a value head. To
make the training process more stable and easier to tune, we separate
the actor network from the critic network, that is, the two networks do
not share parameters. In addition, Generalized Advantage Estimation
(GAE)~\citep{schulman2015high} is adopted in the action advantage
$\hat{A}(s,a)$ estimation process to achieve stable and accurate
advantage estimation.

\subsubsection{The Policy}
By modeling the problem as an MDP, each sub-action policy is
independent of each other, and the policy is expressed in logarithmic
form in the actor-critic algorithm. Therefore, as formulated in
(\ref{log_policy}), the full packing step policy $\pi(a|s)$ is the
product of sub-action policies, where subscripts $s,r,p$ denote box
selection, rotation, and position sub-action, respectively.
\begin{align}
\label{log_policy}
\begin{split}
\log\pi(a|s)&=\log\pi_{s}(a_s|s_s)\pi_{r}(a_r|s_r)\pi_{p}(a_p|s_p) \\
&= \log\pi_{s}(a_s|s_s)\!+\log \pi_{r}(a_r|s_r)\!+\log \pi_{p}(a_p|s_p)
\end{split}
    \raisetag{26pt}
\end{align}

\subsubsection{Training Process}
In every training step, the critic network estimates the state values,
and the actor network performs the three steps described before to get
each sub-action output. Thereafter, one-step parameter update is
performed for both the actor and critic networks. Because our training
data is randomly generated and is inexpensive, the on-policy
actor-critic algorithm is applied.

At the same time, choosing the optimal temperature $\alpha^*$ in
(\ref{reward}) is non-trivial since the magnitude of the reward
differs across tasks and also depends on the policy, which improves over time during training.
So we also use an objective function for $\alpha$ and
tune it automatically, where, like in SAC~\citep{haarnoja2018soft}, entropy is considered a constraint.

\begin{align}
\label{loss_all}
\begin{split}
\mathcal{L} &= \mathcal{L}_\theta + \mathcal{L}_\phi + \mathcal{L}_{\alpha} \\
\mathcal{L}_\theta &= - \hat{A}(s_t,a_t) * \sum_{a \in \mathcal{A}}\log \pi_\theta(a_t|s_t)\\
\mathcal{L}_\phi &= MSE[(V_{\phi}(s_t),\hat{A}(s_t,a_t)+V_{\phi}(s_t)]\\
\mathcal{L}_{\alpha} &= \alpha_t[\bar{\mathcal{H}} - \log \pi_t(a_t|s_t)]
\end{split}
\end{align}

Equation (\ref{loss_all}) formulates the loss function calculation
process, which includes the actor, critic and entropy regularization
temperature loss functions. The logarithm of policy $\log
\pi_\theta(a_t|s_t)$ consists of three sub-action policies as shown in
(\ref{log_policy}).
The actor loss $\mathcal{L}_\theta$ is the advantage multiplying
the policy gradient that encourages actions to achieve higher
accumulate rewards. The critic network is trained through Mean Square Error (MSE) loss $\mathcal{L}_\phi$, in which $V_{\phi}(s_t)$ is the value function estimation. The entropy regularization temperature loss
$\mathcal{L}_{\alpha}$ keeps the policy entropy close to the target
entropy $\bar{\mathcal{H}}$ to balance between exploitation and exploration.

\section{Experiments}
\label{experiments}

We evaluate the RCQL model on 2D and 3D strip
packing problems, for both online and offline versions, with different
packing box numbers and compare the
results with heuristic algorithms and learning-based
methods.

\subsection{Experimental Setup}
\subsubsection{Packing Environment}
In our packing problem environment, the model only needs to generate the
index, rotation and the horizontal coordinates of the packing box at
every packing step. Considering the gravitational property of boxes, the box
should be supported by the bin or other boxes. Therefore, the
environment will automatically drop the box $i$ to the lowest available
position in the bin, which is formulated as $z_i = \max_{b}({z_b+h_b})$.
The support box $b$ satisfies the following constrains:
\begin{equation}
\label{below_constrain}
\begin{aligned}
x_i&<x_b+w_b \\
x_b&<x_i+w_i \\
y_i&<y_b+l_b \\
y_b&<y_i+l_i
\end{aligned}
\end{equation}
in which $x,y,z$ denote the left-back-bottom coordinates of the box, and
$w,l,h$ denote the box width, length, and height, respectively. The equation
(\ref{below_constrain}) ensures that box $b$ supports the upper box $i$.
To avoid the model generating positions outside of the bin, as
formulated in (\ref{force_inside}), the environment forces the
cross-border boxes to the bin border, where $W,L$ denote the bin width
and length, respectively. The number of discrete position slots $n_s$ is
set to 128. We find this to be enough to obtain good packing results
while it would not slow down the training process too much.

\begin{equation}
\label{force_inside}
\begin{aligned}
x_i &= \min(x_i, W-w_i)\\
y_i &= \min(y_i, L-y_i)
\end{aligned}
\end{equation}
\subsubsection{Dataset}
In our packing environment, $N$ boxes are initialized with random width,
length and height. The bin is initialized with a fixed width and length,
and normalized to $-1\sim1$, which can be easily scaled to any bin size.
Unlike previous datasets, due to random sampling, the sizes of boxes here
are strongly heterogeneous, which makes the problem harder. The
agent has to find a solution sequence of candidate boxes using minimum
bin height without any overlapping of boxes. In order to show the
generalization ability of the methods, we evaluate the algorithms on one
plain dataset ($w,l,h \sim U(0,L_b)$) and one hard dataset ($w,l,h \sim
U(0,L_b/4)$), where $L_b$ denotes the length and width of the bin.

There are two reasons for choosing randomly synthesized datasets. First,
reinforcement learning requires lots of data to train. Thus, it is
impractical to use real-world datasets in training. In fact, synthetic
datasets are widely used in neural combinatorial optimization
literature~\citep{kool2018attention,chen2019learning,joshi2019efficient,xing2020graph}.
Second, our dataset is generated by uniform sampling, and we all know
that the uniform distribution has the largest entropy, which means
that it is the hardest one. Therefore, the results on this dataset can
speak for the quality of the method.

\subsubsection{Implementation Details}

We experiment with two sizes of models. Our small models have 3 encoder
layers and a decoder layer with a hidden size $d_h=128$, except the
feedforward ReLU layers, which have 512 units. The large models have 6
encoder layers and 2 decoder layers with a hidden size $d_h=256$, and a
feedforward size of 1024. All models have 8 attention heads in each layer. Each
attention layer applies batch normalization. Compared with the model
with no recurrence, the recurrent model can
increase the context information from $B$ to $L*B$, where $L$ and $B$
are the number of encoder layers
and the length of the recurrent FIFO, respectively. Therefore, we set
$B$ as 20 based on
the dataset and the model size to balance the performance and memory
cost. Eventually, our small model has 2.16M
parameters, while the large one has 4.24M parameters.

We use Adam~\citep{kingma2015adam} with a batch size of 128 and a fixed
learning rate of $10^{-4}$. The discount rate $\gamma$ is set as 0.96.
We train our RCQL model on instances of 200 boxes with 10000 training
steps and then evaluate it with a greedy decoder on various numbers of
boxes to show the scalability of our model. We also adopt gradient
clipping at 5.0 for better
stability. The target entropy $\bar{\mathcal{H}}$ is set as 0.6 for automatic
temperature adjustment. The model is trained on a single GeForce 2080Ti
GPU. It takes about 2 days training for 3D offline cases.\footnote{Implementations
are available at \url{https://github.com/dongdongbh/RCQL}.}

\subsubsection{Baselines}

We compare our method with heuristic
algorithms and learning-based methods. Here we choose the best heuristic
methods that are currently known~\citep{jylanki2010thousand},
MAXRECTS\_BL for offline
cases, and SKYLINE\_BL for online cases. For meta-heuristic algorithms,
a Genetic Algorithm (GA)~\citep{wu2010three} and a Simulated
Annealing (SA)~\citep{rakotonirainy2020improved} one are tested. In GA, the
population size and number of generations are 120 and 200 respectively.
In SA, same as in the original paper, the search is terminated when
5000 search iterations has been carried out.
For learning-based algorithms, the multi-task selected learning
(MTSL)~\citep{duan2019multi} model is evaluated.
For comparison purpose, we implement their model but set the same reward function as ours. To verify the effectiveness of our conditional query mechanism, we
remove the conditional query of our model (No query) and get the box
rotation and position from the box selection decoder. Besides, we also
test the \textit{rollout} baseline~\citep{kool2018attention} with the
REINFORCE algorithm, which was claimed to be computationally more
efficient.

\subsection{Performance Evaluation}

We evaluate previously mentioned algorithms by
the bin gap ratio $\rho = (1-\frac{\sum_{i=1}^n w_i l_i h_i}{WLH_n})\times 100\%$,
which is positively related to the final bin height. The variance of the bin
gap ratio is also evaluated to show the stability of the learning
algorithms. Because other learning-based methods cannot handle large scale packing problems,
the case with 40 boxes is selected for comparison.
Large scale problems are tested in the scalability evaluation part.

\begin{table}[tbhp]
\centering
\caption{Results in 40 boxes cases on hard dataset ($w,l,h \sim U(0,L_b/4)$). The variances of the learning methods are calculated from 4 inference runs, the batch size in each run is 128, and the variances of other methods are obtained from 512 separate runs.}

\resizebox{\columnwidth}{!}{%
\begin{tabular}{cl ccccr cc}
\toprule

{} & \multirow{2}{*}{Method}  & \multicolumn{5}{c}{Offline} & \multicolumn{2}{c}{Online} \\

\cmidrule(lr){3-7}
\cmidrule(l){8-9}
\multicolumn{1}{l}{} & {} &\multicolumn{1}{l}{Worst (\%)} & \multicolumn{1}{l}{Best (\%)} & Average (\%) & Variance & Time (ms) & Average (\%) & Variance \\

\midrule

\multirow{6}{*}{2D}
 & Heuristic  & 26.12 & 6.80 & 15.81 & 0.0018 & 11 & 21.11 & 0.0064 \\
 & GA  & 65.59 & 34.65& 49.70 & 0.0026 & 1386 & - & -\\
 & SA  & 25.94 & 7.39& 15.37 & 0.0017 & 12692 & - & -\\
 & Rollout  & 40.57 & 19.34 & 28.12 & 0.0021 & 439 & -  & -\\
 & No query   & 38.26 & 17.98 & 25.73 & 0.0012 & 598 & 26.46  & 0.0003 \\
 & RCQL  & 27.34 & 9.74 & \textbf{14.56} & 0.0015 & 624  & \textbf{15.57} & 0.0003 \\
 & RCQL large   & 25.03 & 8.87 & \textbf{13.98} & 0.0020 &  978  & \textbf{14.86} & 0.0002\\

\midrule

\multirow{7}{*}{3D}
 & Heuristic  & 63.29 & 34.16 & 46.37 & 0.0023 & 15  & 52.97 & 0.0072 \\
 & GA  & 72.72 & 45.51 & 59.57 & 0.0018 & 4312 & -  & - \\
 & MTSL  & 70.56 & 45.87 & 54.29 & 0.0033 & 1924 & - &- \\
 & Rollout  & 57.67 & 30.97 & 38.09 & 0.0034 & 690 & - & - \\
 & No query & 55.29 & 27.77 & 33.14 & 0.0025 & 765 & 49.83 & 0.0003 \\
 & RCQL  & 52.84 & 28.97 & \textbf{31.25} & 0.0012 & 828 & \textbf{48.37} & 0.0003 \\
 & RCQL large  & 49.23 & 27.12 & \textbf{30.25} & 0.0015 & 1037 & \textbf{46.12} & 0.0002 \\
\bottomrule
\end{tabular}%
}
\label{tbl:gap_ratio}
\end{table}

Table~\ref{tbl:gap_ratio} shows numerical results of 512 instances with
40 boxes for both 2D and 3D cases. Our RCQL model achieves lower bin gap
ratio in both 2D and 3D, online and offline cases. Specifically, in 2D
cases, although the small RCQL model has only 2.49M parameters, it achieves
14.56\% and 15.57\% average gap ratios, in offline and online situations,
respectively. The best heuristic algorithm shows superior
performance to GA and learning-based baselines since it
applies explicit rules that make boxes cling to each other and do not
consider the gravitational property of boxes which makes the solution space
larger than others.
SA achieves a lower gap ratio than the heuristic one since it searches for
better results among heuristic construction solutions. However, SA
requires a lot of iterations to search from scratch, making the time
cost of SA much higher.
While our method already learned the packing pattern after training,
it is faster than these meta-heuristic methods in inference time.
For the 3D cases, learning-based approaches show lower gap ratios, and models
which have the attention mechanism show better result
than the others. The variance of the RCQL model is relatively small,
which means that it can stably learn the general pattern of the packing
problems. Furthermore, larger RCQL models also show slightly better
results than small models.

It is clear that the model with the conditional query mechanism is better
than the no query model, which confirms that the conditional query fills
the information gap between sub-actions and makes the learning algorithm capable
of reasoning for the following sub-actions according to the embedded
previous outputs. Meanwhile, the rollout baseline with REINFORCE
produces similar results to the \textit{no query} model but with higher
variance. Because the rollout method sums up all rewards in every
packing step as the action value, its learning process treats every
packing step equally and back-propagates gradients for all steps
regardless of whether the quality of actions is good or bad.
MTSL shows poor results,
because their setting is only a partially observable MDP.
Because of the information gap between sub-actions,
MTSL also shows very high variance results.

Learning-based methods are slower than the heuristic
algorithms since the computational load of NNs is relatively large. But
it is faster than meta-heuristics, because the searching process of
meta-heuristics is time-consuming.
In addition, the NN model can benefit from batch input, so the time cost
per input instance is much less than the time reported. Due to
recurrence, the RCQL model is slightly slower than the other learning
methods except MTSL in the 40 boxes case. But the baseline models can
not handle large scale problems since the memory cost will be unbearable.
While the time cost of the RCQL model is only linearly proportional to
the box number owing to its recurrent feature. It only takes about 21 seconds
when the number of boxes reaches 1000 and it achieves a 5.64\% lower bin
gap ratio than GA.

\begin{figure}[tbhp]
\centering
    \includegraphics[width=.7\columnwidth]{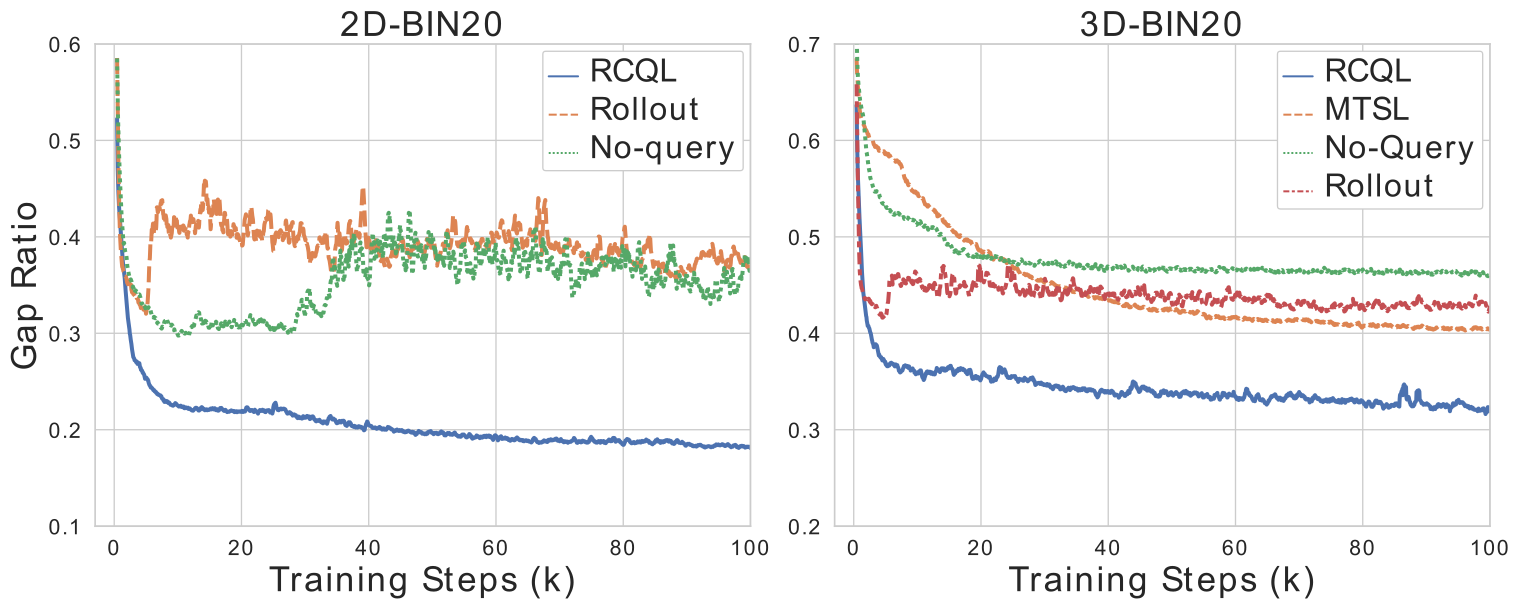}
    \caption{Learning curve of different models. Tested with the 20 boxes case.}
    \label{lc}
\end{figure}

From the learning curve of the training process shown in Fig.~\ref{lc}, the
RCQL model also shows superior stability to the other learning algorithms,
which leads to lower variance as shown in Table~\ref{tbl:gap_ratio}. The
learning curve of the no query model is oscillating as the sampled
results of earlier steps are not passed on to the later steps.
That is, the model can
only estimate the solution that is good overall but not one that fits
the instance with specific input state.
The learning curve of the rollout method shows a big spike at
the beginning of training process because of the sudden baseline update
after the first epoch. In contrast, the RCQL model benefits from
conditional query to construct a strict MDP and has meaningful reward
signals from every box packing step, so it gives a smooth convergence
process.

\subsection{Scalability Evaluation}

In order to evaluate the scalability and generalization ability of these
algorithms, we use different box numbers and different datasets in the
experiments. Because no recurrence model must embed all box states to
infer actions, the memory cost increases quadratically with the the increase of
the box number. When the number exceeds 60, the baseline models all run
out of GPU memory during training time. So we only evaluate these models
for no more than 60 boxes.
Another major drawback of these models is that their structures are
problem size-related. As a result, for each problem instance with a
certain number of boxes, a new model must be trained.
In contrast, only one trained RCQL model is needed to handle various
numbers of boxes; that is, 20 and 1000 boxes cases use the same RCQL
model.

\begin{figure}[tbhp]

\centering
\includegraphics[width=.7\columnwidth]{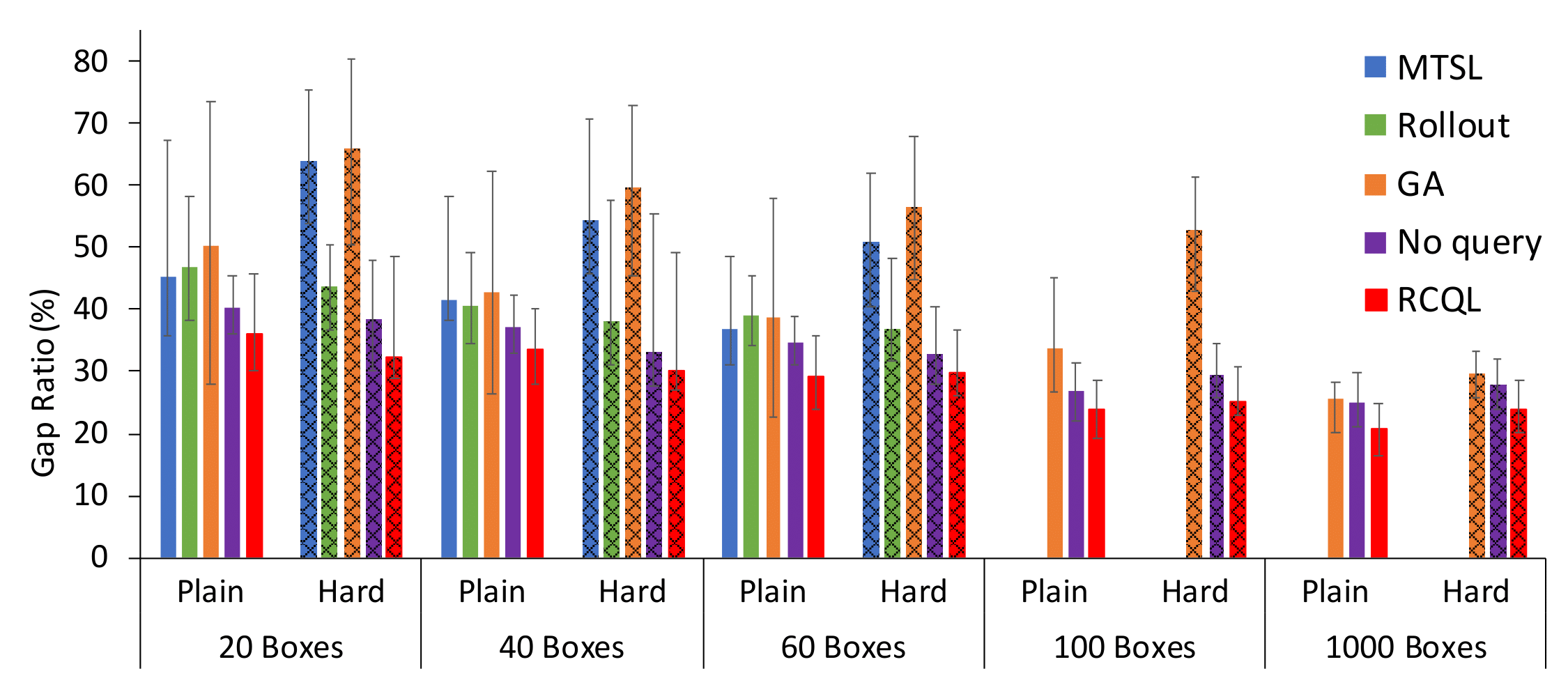}
\caption{Bin gap ratio with different box numbers and datasets in 3D offline cases. Lower is better.}
\label{fig:scale}

\end{figure}

As shown in Fig.~\ref{fig:scale}, the bin gap ratio of RCQL is lower than
all the baseline ones for various datasets.
The overall gap ratio of all methods
decreases as the packing box number increases. The rollout and no query
models show lower gap ratio than the other baselines, which means the
attention models are more efficient in learning the packing pattern.
RCQL achieves the highest space utilization ratio with only one trained
model for all datasets and different box numbers. Although the genetic
algorithm becomes better than MTSL when the box number increases, it
is still inferior to RCQL since it tends to get trapped in local optima.

\subsection{Training/Testing Generalization}

To further study the generalization ability of the RCQL model, five models
are trained through hard datasets with different instances sizes, which
are then
evaluated on testing sets with various instance sizes.

\begin{table}[tbhp]

\centering
\caption{Testing results of various sizes of
RCQL models trained through different instance sizes.}
\begin{tabular}{cllllll}
  \toprule
 \multirow{2}{*}{Train}&\multicolumn{6}{c}{Test} \\
\cmidrule(lr){2-7}
&
  40 &
  100 &
  200 &
  500 &
  800 &
  1000 \\
\midrule
40 &
  32.43 &
  29.27 &
  27.95 &
  25.79 &
  25.56 &
  26.50 \\
100 &
  31.36 &
  \textbf{27.47} &
  \textbf{24.50} &
  23.97 &
  22.90 &
  22.84 \\
200 &
  \textbf{31.25} &
  27.97 &
  24.98 &
  \textbf{22.92} &
  22.18 &
  22.11 \\
500 &
  34.94 &
  28.96 &
  25.45 &
  23.10 &
  \textbf{22.06} &
  \textbf{21.79} \\
800 &
  37.40 &
  32.71 &
  26.88 &
  23.37 &
  22.30 &
  22.48 \\
  \bottomrule
\end{tabular}

\label{tab:tv}
\end{table}

As shown in
Table~\ref{tab:tv}, the bin gap ratios of different models tested on
specific instance sizes come out to be very similar. In some cases,
the model trained with different instance sizes performs even
better than the one trained with the same instance size. This shows
that test results do not rely on the instance sizes of the training sets,
and the model can be satisfactorily generalized by learning the packing pattern.

Generally, when the instance size is larger, the bin gap ratio is lower.
In particular, the model trained with 40-box instances results in
a relatively high gap since it has no chance to learn the generalization
ability of long-term dependency. In contrast, a model trained with too
large an instance size will cause the Bellman
backup~\citep{greene2019reinforcement} in value iteration to be non-trivial,
which will also lead to performance degradation.

\subsection{Going Deeper with Context}

To study the effect of context information, we experiment with different
context sizes of packed boxes $n_p$ and unpacked boxes $n_u$. The results
are shown in Table~\ref{tab:context}, where two small models (3 encoder
layers and 1 decoder layer) are trained in each specific context
setting---one has recurrent connections, the other does not. The training
GPU memory costs of these two models under specific $n_p$ and $n_u$ are
similar, so we only report it in the recurrence case. Dynamic state
updating is employed in both models.

\begin{table}[tbph]
\centering
\caption{Empirical results for context size and recurrence. All models are trained and tested on 3D hard instances with 200 boxes. The training set contains $128*10^4$ instances.}
\begin{tabular}{@{}cccccc@{}}
\toprule
\multirow{2}{*}{$n_u$} &
  \multirow{2}{*}{$n_p$} &
  \multicolumn{2}{c}{Bin Gap Ratio $\rho$ (\%)} &
  \multirow{2}{*}{\begin{tabular}[c]{@{}c@{}}GPU \\ Memory (GB)\end{tabular}} \\ \cmidrule(lr){3-4}
                    &    & W/ Recur    & W/O Recur    &      &      \\ \midrule
\multirow{3}{*}{10} & 5  & 33.91 & 34.66 & 1.758 \\
                    & 10 & 32.83 & 33.97 & 2.246 \\
                    & 20 & 32.58 & 30.61 & 3.184 \\ \midrule
\multirow{3}{*}{20} & 5  & 29.24 & 32.27 & 2.064 \\
                    & 10 & 28.36 & 30.92 & 2.770 \\
                    & 20 & 25.98 & 26.95 & 3.586 \\ \midrule
\multirow{3}{*}{30} & 10 & 26.84 & 31.72 & 2.830 \\
                    & 20 & 26.38 & 28.34 & 4.386 \\
                    & 30 & \textbf{24.25} & 26.56 & 5.172 \\ \bottomrule
\end{tabular}
\label{tab:context}
\end{table}

As shown in Table~\ref{tab:context}, larger context sizes lead to lower
bin gap ratio, which attests to the fact that more state information helps the
learning of the effective packing strategies. If the recurrent connection is
removed, the bin gap volume increases. The smaller the packed state
context size $n_p$, the more evident would be the improvement of
adding the recurrent connection.
This is because a small $n_p$ would lose the long range context,
which is important for constructing the MDP and the learning process.
The recurrent
connection extends the context length to $L*n_p$, where $L$ is the layer
number of the recurrent encoder, which is 3 here. Specifically,
the recurrence model bin gap ratio $\rho_{n_u=30,n_p=10}^{r}$ is
26.84, which is close to the gap ratio of the no-recurrence model
$\rho_{n_u=30,n_p=30}^{nr}$ since they have same context size. Therefore,
models with recurrent connections can achieve similar performance with
lower memory costs.

Although a larger context size achieves better bin space utilization, the
increase in memory overhead grows quadratically with the sequence length,
as mentioned in Section~\ref{recurrent}. When the context size becomes
larger, the performance improvement decreases as the unit context size
increases, because the context with a long distance has little effect on
the next packing action. Too large a context size also makes the training
time unacceptable; in our tests, when both $n_u$ and $n_p$ were 50,
it took several weeks to train and the training memory cost was about 11
GB. As a result, in previous evaluations, both $n_u$ and $n_p$ were set to
20 in order to strike a balance between optimality and training cost.

\section{Conclusion}

\label{conclusion}
In this paper, we propose a recurrent conditional query learning method
to solve packing problems. The RCQL model can be generalized to tackle
any problem with mutually conditioned actions and can solve large scale
strip packing problems with a small number of parameters. Numerical
results show that the RCQL performs better than state-of-the-art methods
for both 2D and 3D strip packing problems.

The success of RCQL proves that the conditional query
mechanism is an efficient method for solving problems with parameterized
action space. Besides, incorporating recurrent feature in NCO is an
efficient way to handle large scale problems in operations research.
The recurrent feature indeed has the potential to improve
the scalability of NCO methods applied in other optimization problems
such as routing.

\bibliographystyle{unsrtnat}
\bibliography{ref}  

\end{document}